%% file: acl_latex.tex
\pdfoutput=1
\documentclass[11pt]{article}

\usepackage[final]{acl}

\usepackage{times}
\usepackage{latexsym}

\usepackage[T1]{fontenc}
\usepackage[utf8]{inputenc}

\usepackage{amssymb}
\usepackage{microtype}
\usepackage{inconsolata}
\usepackage{subcaption}
\usepackage{array}
\usepackage{hyperref}
\usepackage{inconsolata}
\usepackage{url}
\usepackage{amsmath}

\usepackage{amsfonts}
\usepackage{helvet}
\usepackage{courier}
\usepackage{url}
\usepackage{float,color}
\usepackage{colortbl}
\usepackage{bm} 

\usepackage[ruled, vlined, linesnumbered]{algorithm2e}

\usepackage{float}
\usepackage{pdfpages}
\usepackage{changes}
\usepackage{epsfig}
\usepackage{graphicx}
\usepackage{stfloats}
\usepackage{url}
\usepackage{diagbox}
\usepackage{epstopdf}
\usepackage{multicol}
\usepackage{makecell}
\usepackage{longtable}
\usepackage{dsfont,amsfonts,color}
\usepackage{multirow}
\usepackage{booktabs}
\usepackage{svg}
\usepackage{xr}
\usepackage{tablefootnote}
\usepackage{tabularx}
\usepackage{svg}

\def\0{{\bf 0}}
\def\1{{\bf 1}}

\graphicspath{{img/}}

\usepackage{inconsolata}

\usepackage{graphicx}

\title{MixKVQ: Query-Aware Mixed-Precision KV Cache Quantization for Long-Context Reasoning}
\author{
 \textbf{Tao Zhang\textsuperscript{1}},
 \textbf{Ziqian Zeng\textsuperscript{1}\thanks{Corresponding author}},
 \textbf{Hao Peng\textsuperscript{2}},
 \textbf{Huiping Zhuang\textsuperscript{1}},
 \textbf{Cen Chen\textsuperscript{1,3}}
\\
 \textsuperscript{1}South China University of Technology, China,\\
 \textsuperscript{2}Beihang University, China,\\
 \textsuperscript{3}Pazhou Laboratory, China, 
\\
 \small{
   \textbf{Correspondence:} \href{mailto:email@domain}{zqzeng@scut.edu.cn}
 }
}

\newcommand{\mX}{\mathbf{X}}

\begin{document}
\maketitle
\input{latex/sections/0_abstract}
\input{latex/sections/1_introduction}
\input{latex/sections/2_related_work}

\input{latex/sections/3_preliminary}
\input{latex/sections/4_method}

\input{latex/sections/5_experiment}
\input{latex/sections/6_conclusion}
\input{latex/sections/7_limitationsAndRisks}
\bibliography{custom}
\clearpage
\appendix
\input{latex/sections/8_appendix}

\end{document}

%% file: latex/sections/0_abstract.tex
\begin{abstract}
Long Chain-of-Thought (CoT) reasoning has significantly advanced the capabilities of Large Language Models (LLMs), but this progress is accompanied by substantial memory and latency overhead from the extensive Key-Value (KV) cache.
Although KV cache quantization is a promising compression technique, existing low-bit quantization methods often exhibit severe performance degradation on complex reasoning tasks.
Fixed-precision quantization struggles to handle outlier channels in the key cache, while current mixed-precision strategies fail to accurately identify components requiring high-precision representation.
We find that an effective low-bit KV cache quantization strategy must consider two factors: a key channel's intrinsic quantization difficulty and its relevance to the query.
Based on this insight, we propose \textbf{MixKVQ}, a novel plug-and-play method that introduces a lightweight, query-aware algorithm to identify and preserve critical key channels that need higher precision, while applying per-token quantization for value cache.
Experiments on complex reasoning datasets demonstrate that our approach significantly outperforms existing low-bit methods, achieving performance comparable to a full-precision baseline at a substantially reduced memory footprint.
\end{abstract}

%% file: latex/sections/1_introduction.tex
\section{Introduction}

\begin{figure}[t]
\begin{center}
\includegraphics[width=\linewidth]{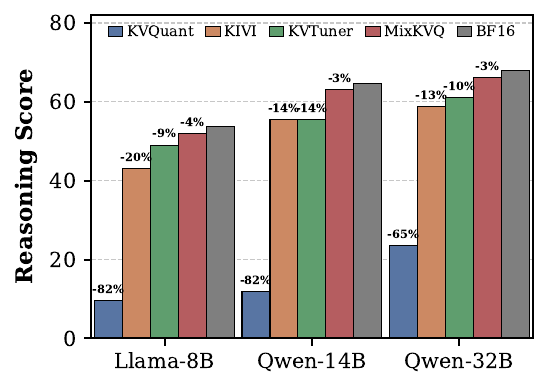}
\end{center}
\vspace{-1.0em}
\caption{Complex reasoning performance of different 2-bit KV cache quantization. Reasoning score is the average accuracy of AIME 2024-2025, MATH 500, GPQA, and LiveCodeBench. }
\label{fig:reason_drop}
\vspace{-1.5em}
\end{figure}

Recent pioneering Large Language Models (LLMs), such as OpenAI-o1 \cite{openai-o1}, DeepSeek-R1 \cite{DeepSeekR1}, and Gemini 2.5 Pro \cite{Gemini15}, have demonstrated remarkable reasoning abilities by generating long Chain-of-Thoughts (CoT).
To achieve superior performance, these models are trained to generate up to 128K tokens with multiple complex rationales from different perspectives. 
However, the auto-regressive nature of LLM inference introduces substantial memory overhead, stemming from the storage demands of the progressively growing Key-Value (KV) cache  \cite{MLSYS2023_c4be71ab}. 
As generating tokens sequentially, LLMs must retain all previous key and value activations in memory to compute attention scores. 
Consequently, the memory footprint of KV cache grows linearly with the sequence length.
For instance, a 32B Qwen2.5 model \cite{Qwen25}, with a batch size of 64 and a sequence length of 32,768 tokens (both prefilled and decoded), requires approximately 512 GB of GPU memory. 
This requirement is 8.59× greater than the memory used to store the model’s weights.
Furthermore, the attention mechanism must repeatedly access this expanding KV cache at each decoding step, creating significant memory bandwidth pressure. 
This recurrent data movement leads to bandwidth saturation, establishing a critical memory-bound bottleneck \citep{Orca, Yuan2024LLMIU} that severely constrains inference throughput.

More recently, KV cache quantization has emerged as a promising solution.
It converts the cache from high-precision floating-point formats to low bit-width integer representations. 
Channel-wise Key and token-wise Value asymmetric quantization has garnered much attention for its high accuracy and tuning-free nature \cite{kivi,kvquant,liu2025quantization}.

However, as demonstrated in Figure \ref{fig:reason_drop}, existing quantization methods lead to dramatic degradation on complex reasoning tasks when pushed to extreme bit-widths such as 2-bit.
Their failures stem from following primary limitations.
First, fixed-precision methods \citep{kvquant, kivi, skvq, sinkq} apply a uniform bit-width across the entire KV cache. 
This approach fails to accommodate channels with extreme outliers, consequently leading to significant accuracy degradation.
Second, existing mixed-precision methods \citep{KVTuner, mixq, pmkvq} typically allocate bit-widths based on quantization errors in the key or value caches. 
Under strict memory constraints, such heuristics allocate higher bit-widths to KV cache segments with larger quantization errors but over-compress those with smaller ones. 
Yet, these low-error segments may still demand higher precision, resulting in significant degradation on complex reasoning tasks.

The goal of KV cache quantization is a dual objective: it must drastically reduce memory usage without compromising the fidelity of attention computations. 
However, existing methods often treat minimizing quantization error of key and value cache as a direct proxy for preserving fidelity. 
This strategy is flawed, as it may allocate higher bit-width to key channels that exhibit large numerical errors but have a negligible impact on the attention computation. 
Such misallocation yields negligible gains in attention fidelity while inefficiently consuming the limited bit-budget.
Our analysis shows that a key channel with large-magnitude activations may not be influential if its corresponding query activations are small. 
Preserving such channels offers diminishing returns and is an inefficient use of the limited bit budget. 
Therefore, an effective quantization strategy must consider two factors: a channel's intrinsic quantization difficulty and its relevance to the query.

Based on this insight, we propose \textbf{MixKVQ}, a novel plug-and-play method that introduces a lightweight, hybrid query-aware heuristic to identify and preserve critical channels in higher precision. 
This heuristic dynamically estimates the potential fidelity impact of quantizing each key channel by combining its intrinsic quantization difficulty with its relevance to the query. 
Channels with a high estimated impact are preserved in higher precision.
MixKVQ quantizes key cache per-channel with mixed precision and quantizes value cache per-token.
To evaluate the effectiveness of our method, we conduct experiments on various models including Llama-3.1~\citep{Llama3}, and Qwen2.5~\citep{Qwen25} family.
Compared with the previous quantization method, our approach can achieve superior performance under different average bit widths in mathematical reasoning tasks \cite{AIME2024, AIME2025, math500, gpqa}, as shown in Figure \ref{fig:reason_drop}. 
Our contributions are summarized as follows:
\begin{itemize}
    \item 
    We find that for effective low-bit KV cache quantization, the precision allocated to a key channel must be determined by two factors: its intrinsic quantization difficulty and its dynamic relevance to the query. 
    \item 
    Building on this insight, we propose \textbf{MixKVQ}, a novel plug-and-play method for extreme low-bit KV cache quantization.
    MixKVQ quantizes key cache per-channel with mixed precision and quantizes value cache per-token.
    \item 
    Experiments on mathematical reasoning benchmarks demonstrate that MixKVQ consistently outperforms existing methods across various compression rates, achieving performance on par with the full-precision baseline. 
\end{itemize}

%% file: latex/sections/2_related_work.tex
\section{Related Work}
\noindent \textbf{KV cache quantization.}
Recent advances in KV cache quantization focus on low-bit compression and asymmetry-aware optimization.
Representative techniques include rotation-based transformations to handle outliers \cite{QuaRot, RotateKV, skvq}, tuning-free asymmetric bit allocation \cite{kivi, sinkq, AsymKV}, and mixed-precision adaptation \cite{KVTuner, kvquant}. 
However, most existing quantization methods are proposed for non-reasoning task.
Intuitively, KV cache quantization may pose greater risks in this context, as reasoning models typically involve long Chain-of-Thought (CoT) outputs \cite{COT}, which are more prone to quantization error accumulation over the sequence.

\noindent \textbf{KV cache eviction.}
Researchers have proposed other several KV cache compression techniques.
One direction is KV cache eviction methods \cite{efficient,H2O,LM-Infinite,pyramidkv,d2o,identifycriticalkvcache,keepkv,cake}improve computational efficiency during model inference by retaining critical KV entries and discarding non-essential ones. 
However, eviction operations can lead to irreversible information loss, and some methods rely on predefined strategies, struggling to fully capture the dynamic characteristics of attention mechanisms.

\noindent \textbf{KV cache management.}
Another strategy is KV cache management, which is an optimization technique to accelerate LLM inference by improving memory utilization.
KV cache management methods mainly include prefilling-decoding disaggregation \cite{mooncake}, low-rank decomposition \cite{Palu,LoRC,Eigen_Attention,xKV}, offloading \cite{InfiniGen,ren2025characterizing}, prefetching \cite{PRESERVE,dong2025accelerating}, and retrieval \cite{razorattention,ReKV}.
It is worth noting that our fine-grained numerical compression is orthogonal to these methods. 
For instance, it can be combined with structural compression from low-rank methods, or applied within active pages managed by retrieval systems to optimize data representation.

%% file: latex/sections/3_preliminary.tex
\section{Preliminaries}
\label{sec:preliminaries}

\subsection{Transformer and the KV Cache}
The core of a Transformer layer is the self-attention mechanism. The output $\boldsymbol{o}_i^l$ for a query vector $\boldsymbol{Q}_i^l$ at layer $l$ is computed by attending to a sequence of key and value vectors:
\begin{equation}\label{eq:attention_score}
\boldsymbol{a}_i^l = \text{softmax}\left(\frac{\boldsymbol{Q}_i^l (\boldsymbol{K}^l)^\top}{\sqrt{D}}\right), \quad \boldsymbol{o}_i^l = \boldsymbol{a}_i^l \boldsymbol{V}^l,
\end{equation}
where $D$ is the model hidden size, $\boldsymbol{a}_i^l$ denotes the attention weight vector for the $i$-th token at layer $l$. 
During autoregressive generation, the matrices $\boldsymbol{K}^l$ and $\boldsymbol{V}^l$ contain the key and value vectors from all previous timesteps. 
To avoid costly recomputation, these matrices are stored in a \textbf{KV cache}, which grows linearly with the sequence length.

\subsection{KV Cache Quantization}
\label{sec:quant}
The memory footprint of the KV cache can become an inference bottleneck. 
Quantization addresses this by storing the cache in a low-bit format. 
We employ a $B$-bit asymmetric quantization scheme, which approximates a tensor $\boldsymbol{X}$ with its dequantized representation $\tilde{\boldsymbol{X}}$:
\begin{equation}
\label{eq:quant}
Q(\boldsymbol{X}) = \text{round}\left( \frac{\boldsymbol{X} - \boldsymbol{z}}{\boldsymbol{s}} \right),
\end{equation}
\begin{equation}
\label{eq:dequant}
\tilde{\boldsymbol{X}} =  Q(\boldsymbol{X})\cdot \boldsymbol{s} + \boldsymbol{z}.
\end{equation}
The parameters are a zero-point $\boldsymbol{z} = \min(\boldsymbol{X})$ and a scaling factor $\boldsymbol{s} = (\max(\boldsymbol{X}) - \min(\boldsymbol{X}))/(2^B - 1)$. 
The scaling factor $\boldsymbol{s}$, which is determined by the dynamic range of the tensor, serves as a direct proxy for its quantization sensitivity.
A larger $s$ implies that a wider value distribution is compressed into the $2^B$ discrete levels, leading to coarser granularity.
Specifically, the rounding operation in Eq.~\ref{eq:quant} introduces a quantization error $|x_i - \tilde{x}_i|$ for each $x_i \in \boldsymbol{X}$. 
This error is mathematically bounded by $|x_i - \tilde{x}_i| \le s / 2$ (see proof in Appendix \ref{app:quant_error_proof}).
A single outlier can inflate $s$, significantly increasing this error bound and degrading the approximation quality for all other elements.

%% file: latex/sections/4_method.tex
\section {Method}
\label{method}

\begin{figure}[t!]
    \begin{center}
    \includegraphics[width=\linewidth]{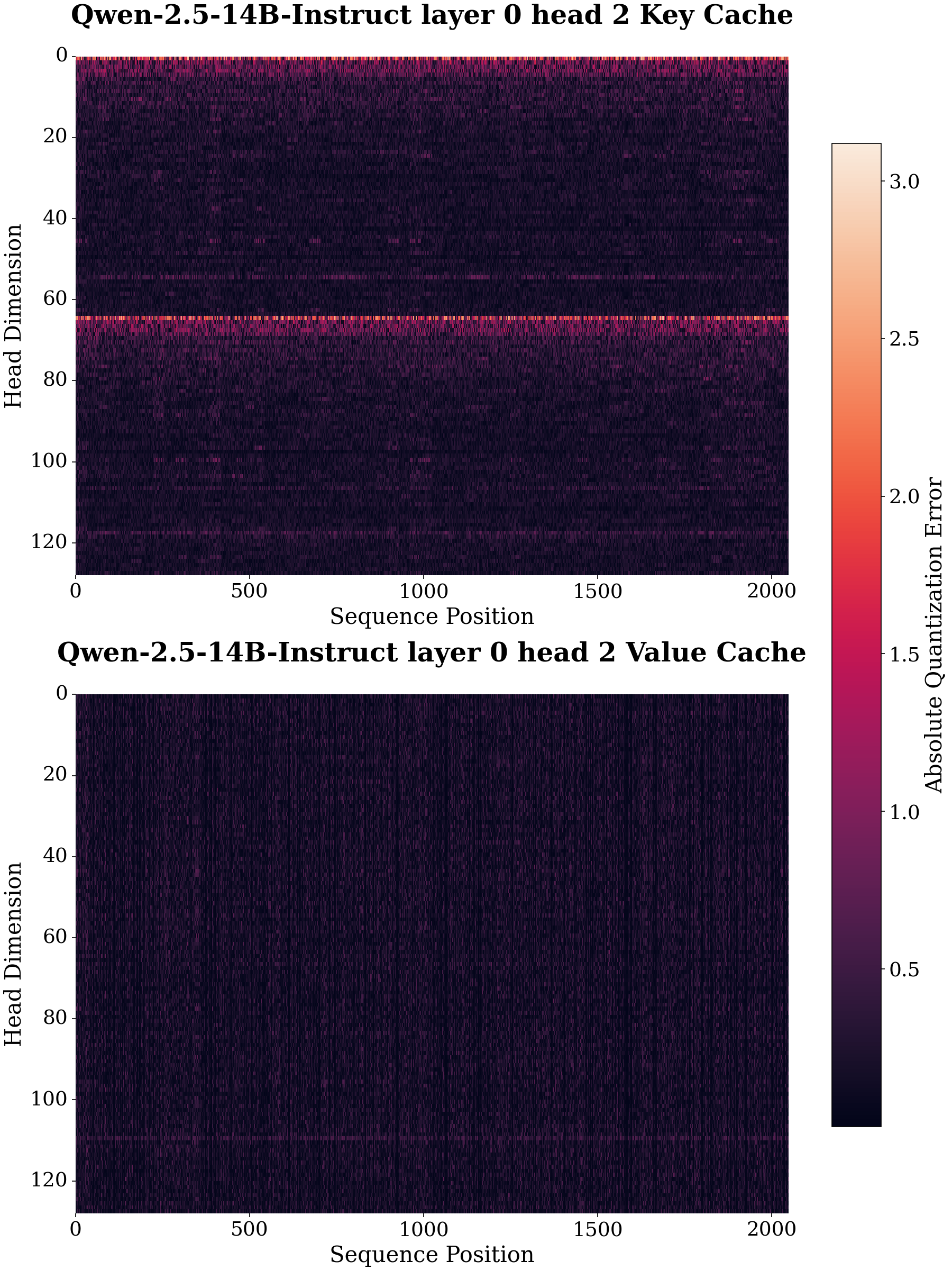}
    \end{center}
    \caption{Absolute quantization error of key and value cache for Qwen-2.5-14B-Instruct model.
    }
    \label{fig:error_heatmap}
    \vspace{-1.0em}
\end{figure}

In long-context or batched inferences, the primary bottlenecks in memory and speed arise from storing and loading the KV cache \citep{Yuan2024LLMIU}.
An effective way to alleviate this problem is to reduce the total number of bytes occupied by the KV cache, specifically through quantization.
However, existing quantization methods leads to dramatic degradation on complex reasoning tasks when pushed to extreme bit-width, detailed in Section \ref{sec: motivation}.
To address this, we introduce MixKVQ, a novel method that introduces a lightweight, hybrid query-aware heuristic to preserve critical channels with higher precision, detailed in Section \ref{sec: MixKVQ design}.

\begin{table*}[t!]
\centering
\caption{
    The error accumulation caused by low-bit KV cache quantization potentially leads to wrong responses in mathematical reasoning tasks.
}
\label{tab:math_reasoning_failure}
\newcommand{\correctbox}[1]{\colorbox{green!25}{\small$#1$}}
\newcommand{\errorbox}[1]{\colorbox{red!25}{\small$#1$}}
\begin{tabularx}{\textwidth}{l X}
\toprule
\multicolumn{2}{p{\dimexpr\textwidth-2\tabcolsep\relax}}{
    \small \itshape\textbf{Question:} Let $\triangle ABC$ have circumcenter $O$ and incenter $I$ with $\overline{IA}\perp\overline{OI}$, circumradius $13$, and inradius $6$. Find $AB\cdot AC$.
} \\
\midrule
BF16 & 
    \small Given triangle \( \triangle ABC \) with circumcenter \( O \) and incenter \( I \), where \( \overline{IA} \perp \overline{OI} \), circumradius \( R = 13 \), and inradius \( r = 6 \). We need to find \( AB \cdot AC \). $\cdots$ Thus, the product \( AB \cdot AC \) is \correctbox{468}. \\
\addlinespace

MixKVQ & 
    \small Given triangle \( \triangle ABC \) with circumcenter \( O \) and incenter \( I \), where \( \overline{IA} \perp \overline{OI} \), circumradius \( R = 13 \), and inradius \( r = 6 \). We need to find \( AB \cdot AC \). $\cdots$ Thus, the product \( AB \cdot AC \) is \correctbox{468}. \\
\addlinespace

KIVI-4bit & 
    \small Given triangle \( \triangle ABC \) with circumcenter \( O \) and incenter \( I \), where \( \overline{IA} \perp \overline{OI} \), circumradius \( 13 \), and inradius \( 6 \). We need to find \( AB \cdot AC \). $\cdots$ So, $480 = x^2 - 2y - (14/13)y = x^2 - (2 + 14/13)y = x^2 - \correctbox{(40/13)y}$. $\cdots$ Thus, the product \( AB \cdot AC \) is \correctbox{468}. \\
\midrule

KIVI-2bit & 
    \small Given triangle \( \triangle ABC \) with circumcenter \( O \) and incenter \( I \), where \( \overline{IA} \perp \overline{OI} \), circumradius \( R = 13 \), and inradius \( r = 6 \). We need to find the product \( AB \cdot AC \). $\cdots$ So, $480 = x^2 - 2y - (14/13)y = x^2 - (2 + 14/13)y = x^2 - \errorbox{(30/13)y}$. $\cdots$ Thus, the product \( AB \cdot AC \) is \errorbox{429}. \\
\addlinespace

KVTuner & 
    \small Given triangle \( \triangle ABC \) with circumcenter \( O \) and incenter \( I \), where \( \overline{IA} \perp \overline{OI} \), circumradius \( R = 13 \), and inradius \( r = 6 \). We need to find the product \( AB \cdot AC \). $\cdots$ So, $x^2 - (2 + 14/13)y = 480$. Which is $x^2 - \errorbox{(30/13)y} = 480$. $\cdots$ Thus, the product \( AB \cdot AC \) is \errorbox{429}. \\
\bottomrule
\end{tabularx}
\vspace{-0.5em}
\end{table*}

\begin{table}[t]
\centering
\small
\caption{The results of simulated KV cache quantization with various configurations. 
All quantization methods are applied at 2-bit precision.}
\label{tab:scal_and_query}
\resizebox{\linewidth}{!}{
\begin{tabular}{llcc}
\toprule
               &                 & \multicolumn{2}{c}{\textbf{Perplexity $\downarrow$}}\\ \cline{3-4}
\textbf{Model} & \textbf{Method} & WikiText2 & C4 \\
\midrule
\multirow{5}{*}{Qwen2.5-7B-Instruct} & BF16 & 6.46 & 2.51\\
& KIVI-KV4 & 6.75 & 2.74\\
& KIVI-K4V2 & 6.81 & 2.77\\
& KIVI-K2V4 & 8.13 & 3.15\\
& KIVI-KV2 & 8.87 & 3.24\\
\bottomrule
\end{tabular}}
\end{table}

\subsection{Quantization Error Analysis}
\label{sec: motivation}
In autoregressive LLMs, KV cache quantization error accumulates across two dimensions: model depth (layer-to-layer) and sequence length (token-to-token). 
While the quantization error for a single token or layer may be negligible, its cumulative effect across thousands of tokens can be substantial \cite{KVTuner}. 
This leads to phenomena like token flipping and cascading generation errors, analogous to issues in model weight quantization \citep{improving-conversational}. 
As shown in Table \ref{tab:math_reasoning_failure}, this error accumulation is particularly detrimental in mathematical reasoning task, where the corruption of a single critical value can invalidate an entire logical chain, squandering significant computational resources.

\noindent \textbf{Key Cache is Generally More Important}.
Figure \ref{fig:error_heatmap} visualizes the absolute quantization error in 2 bit of the key and value cache for the Qwen-2.5-14B-Instruct model (layer 0, head 2). 
Notably, certain channels in the key cache exhibit significantly larger quantization errors. 
In contrast, the error distribution for the value cache is more uniform and lacks distinct outliers. 
As shown in Table \ref{tab:scal_and_query}, our experimental results confirm that preserving higher precision for the key cache is more critical for maintaining model performance. 
This finding aligns with prior work \cite{kivi,keyformer,KVTuner}. 
Therefore, the core challenge is to develop a precision allocation strategy for the key cache that minimizes attention score loss under a constrained memory budget.

\begin{figure*}[t!]
\begin{center}
    \includegraphics[width=0.97\linewidth]{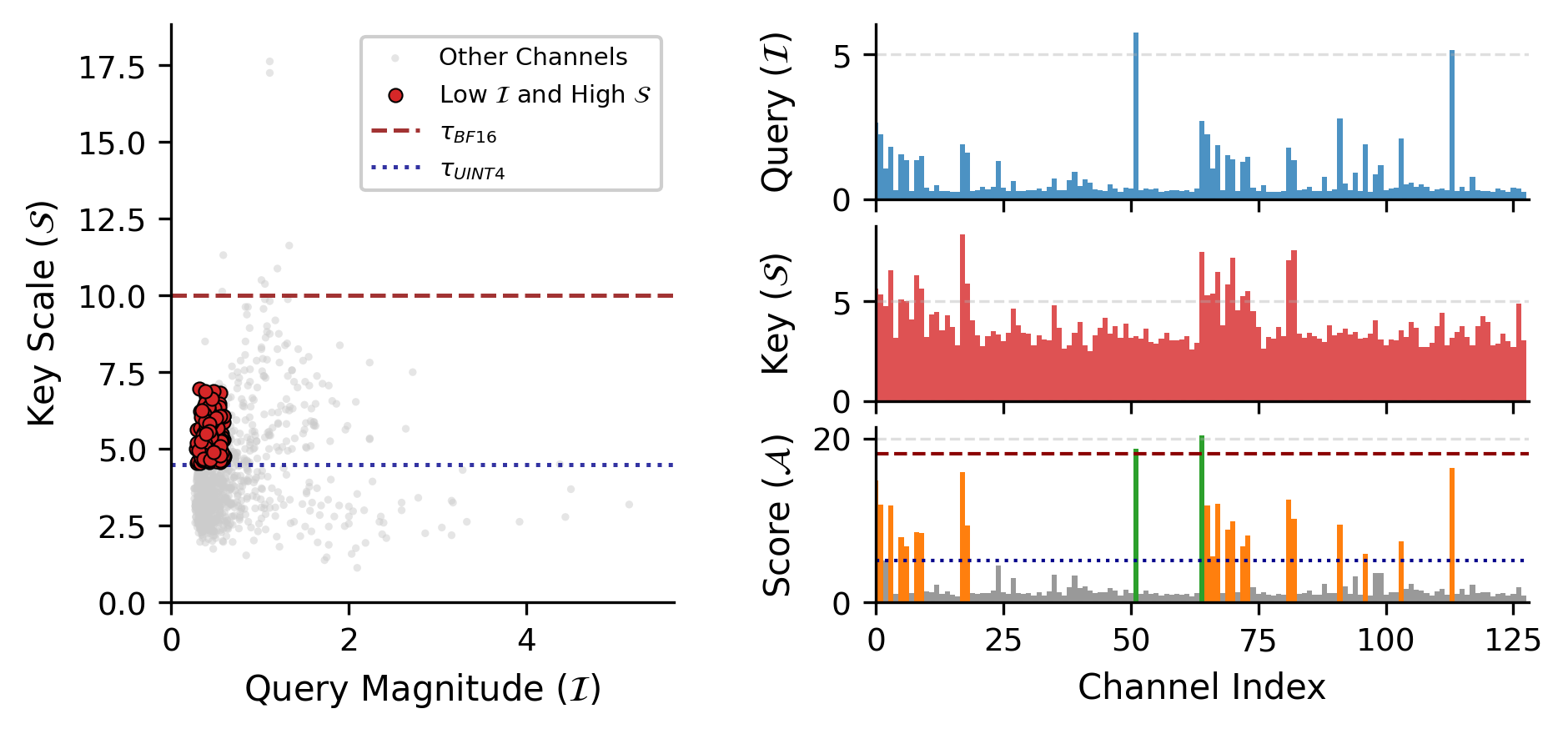}
    \end{center}
    \vspace{-1.0em}
    \caption{
        Analysis of Key channel properties on Qwen-2.5-14B-Instruct. 
        Scatter plot of Query magnitude ($\mathcal{I}$) versus Key scale ($\mathcal{S}$) in Layer 0. 
        Here, $\mathcal{I}$ denotes the average activation intensity of the Query vectors and reflects each channel’s contribution to the attention scores. 
        Traditional methods assign high bit-widths to channels with high $\mathcal{S}$ but low $\mathcal{I}$ (shown as blue dots) due to their large $\mathcal{S}$ values; 
        however, these channels are in fact non-crucial for preserving the fidelity of attention score computation.
        (b) Channel statistics for Head 4. 
        In contrast to the Key scales ($\mathcal{S}$), which are densely clustered and offer limited discriminative capability, the combined salience score ($\mathcal{A}=\mathcal{I}\cdot\mathcal{S}$) effectively isolates critical channels that require high-precision retention. 
        Bar colors denote the adaptive precision levels: green (BF16), orange (INT4), and grey (INT2).
    }
    \vspace{-1.0em}
    \label{fig:key_quant_analysis}
\end{figure*}

\noindent \textbf{Fixed-Precision Methods Struggle with Outliers.}
Existing fixed-precision methods \cite{kivi, kvquant, skvq}, perform poorly at low bit-widths like 2-bit. 
As illustrated in Figure \ref{fig:error_heatmap}, while these methods can adequately compress the value cache, they struggle to handle key cache channels with significant outliers. 
This results in large quantization errors and critical failures on reasoning tasks.

\noindent \textbf{Suboptimal bit-widths Allocation in Existing Mixed-Precision Methods.} 
Current layer-wise and channel-wise approaches \cite{KVTuner,mixq,pmkvq} exhibits shortcomings in bit-widths allocation. 
Layer-wise methods, due to their static nature, are prone to over-aggressive quantization of sensitive layers (see Appendix~\ref{app:kvtuner_failure}). 
Channel-wise heuristics assign higher bit-widths to Key channels with larger scales.
However, Figure~\ref{fig:key_quant_analysis}(a) shows that the magnitude of Query activations exhibits little correlation with Key scales, with a Pearson correlation coefficient of merely 0.16. 
Since high Query activation indicates greater importance for preserving attention fidelity, methods that rely solely on scaling factors may preserve Key channels in higher precision (shown as red dots) that are in fact non-critical to accurate attention score computation. 
Moreover, Figure~\ref{fig:key_quant_analysis}(b) highlights the limited discriminative capability of scales alone: the distribution of $\mathcal{S}$ is highly concentrated, with 80\% of values in Head~0 clustered within the narrow range [2.80, 4.46], making it difficult to differentiate truly critical channels.
Together, these observations demonstrate that $\mathcal{S}$ alone is insufficient for identifying salient channels.

The goal of KV cache quantization is to reduce the KV cache memory footprint while preserving the fidelity of the attention computation.
Our analysis posits that \textbf{a key channel with large-magnitude activations may NOT be critical if its corresponding query activations are small}. 
Preserving such channels offers diminishing returns and is an inefficient use of the limited bit budget.
This motivates the need for a more informative metric that incorporates Query information to enable effective precision allocation, particularly in low-bit regimes.

\begin{figure*}[t!]
\begin{center}
    \includegraphics[width=1.0\textwidth]{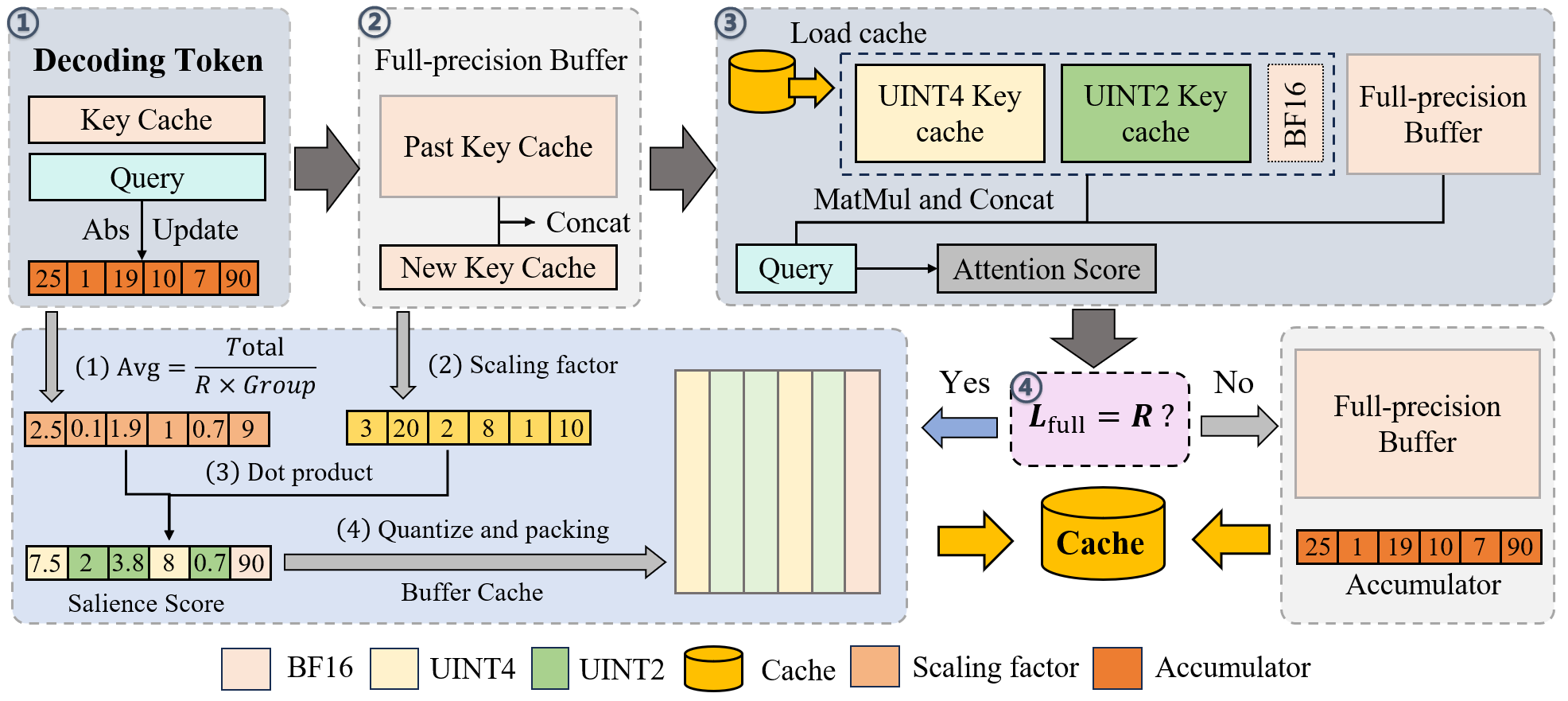}
    \end{center}
    \vspace{-1.0em}
    \caption{
    Workflow of MixKVQ.
    Keys and Values undergo per-channel and per-token quantization, respectively, managed by a full-precision buffer of size $R$.
    For Key cache, salience Score $\mathcal{A}_d$ are updated every $R$ tokens within the window, while Value cache utilize uniform 2-bit quantization.
    }
    \label{fig:method}
    \vspace{-1.0em}
\end{figure*}

\subsection{MixKVQ}
\label{sec: MixKVQ design}
Based on this insight, we propose \textbf{MixKVQ}, a novel method that introduces a lightweight query-aware heuristic to identify and preserve critical channels in higher precision. 
Given that the value cache can be adequately compressed, the main objective is to mitigate the quantization error introduced in the \textbf{attention scores}. 

Let $\mathbf{Q} \in \mathbb{R}^{L_q \times D}$ and $\mathbf{K} \in \mathbb{R}^{L_k \times D}$ denote the query and key matrices, respectively, where $D$ is the hidden dimension.
When $\mathbf{K}$ is quantized to $\tilde{\mathbf{K}}$, the error in the pre-softmax attention scores, denoted as $\mathbf{E}_{attn}$, is defined as:
\begin{equation}
\mathbf{E}_{attn} = \mathbf{Q}(\mathbf{K} - \tilde{\mathbf{K}})^T.
\end{equation}

Consider the error term for the $i$-th query token and $j$-th key token, denoted as $E_{i,j}$. 
Let $\epsilon_{j,d} = \mathbf{K}_{j,d} - \tilde{\mathbf{K}}_{j,d}$ represent the quantization noise for the key at token $j$ and channel $d$. 
The logit error $E_{i,j}$ aggregates the error contributions across the channel dimension:
\begin{equation}
E_{i,j} = \sum_{d=1}^{D} \mathbf{Q}_{i,d} \cdot \epsilon_{j,d}.
\end{equation}
To minimize the impact on the attention mechanism, we aim to identify and preserve channels with the highest expected magnitude of error contribution, $\mathbb{E}[|\mathbf{Q}_{i,d} \cdot \epsilon_{j,d}|]$.
We approximate this expectation as the product of their individual expected magnitudes: $\mathbb{E}[|\mathbf{Q}_{i,d}|] \cdot \mathbb{E}[|\epsilon_{j,d}|]$.
We operationalize this via a heuristic metric, the \textbf{Salience Score} ($A_d$), derived from two low-cost statistics:

\paragraph{Importance Score ($\mathcal{I}_d$):} 
We estimate $\mathbb{E}[|\mathbf{Q}_{i,d}|]$ using the average absolute magnitude of the query channel $d$ over the sequence length $L_q$:
\begin{equation}
\label{equ:importance_score}
\mathcal{I}_d = \frac{1}{L_q} \sum_{i=1}^{L_q} |\mathbf{Q}_{i,d}|.
\end{equation}

\paragraph{Sensitivity Score ($\mathcal{S}_d$):} 
For a uniform quantizer with bit-width $B$, the quantization error $\epsilon_{j,d}$ is bounded by half the scaling factor (i.e., $|\epsilon_{j,d}| \le \frac{\mathcal{S}_d}{2}$). 
Thus, We estimate $\mathbb{E}[|\epsilon_{j,d}|]$ using the scaling factor $\mathcal{S}_d$ for channel $d$:
\begin{equation}
\label{equ:sensitivity_score}
\mathcal{S}_d = \frac{\max(\mathbf{k}_d) - \min(\mathbf{k}_d)}{2^B - 1},
\end{equation}
where $\mathbf{k}_d$ represents the vector of key values in channel $d$ across the current tokens.

Finally, the \textbf{Salience Score} $\mathcal{A}_d$ is defined as the product of these components:
\begin{equation}
\label{equ:salience_score}
\mathcal{A}_d = \mathcal{I}_d \cdot \mathcal{S}_d.
\end{equation}
Channels with high $\mathcal{A}_d$ are deemed \textbf{critical channels}, as they exhibit both high query relevance and high quantization sensitivity, thereby warranting higher-precision retention.

We designed a three-tiered mixed-precision strategy. 
While prior studies indicate that 4-bit quantization offers a favorable trade-off between model performance and memory footprint~\cite{kivi, qserve, liu2025quantization}, we further optimize this balance by allocating bit-widths based on channel salience. 
To preserve generation quality, the most critical channels are retained in full precision (BF16). 
Simultaneously, to maximize compression, we quantize moderately critical channels to UINT4 and aggressively compress non-critical channels to UINT2. 
Formally, we introduce two thresholds, $\tau_{\text{BF16}}$ and $\tau_{\text{UINT4}}$, applied to the salience score $\mathcal{A}_d$:
\begin{itemize}
    \item \textbf{High-Precision (BF16):} Channels with $\mathcal{A}_d > \tau_{\text{BF16}}$ are identified as highly critical and retained in BF16.
    \item \textbf{Medium-Precision (UINT4):} Channels satisfying $\tau_{\text{UINT4}} < \mathcal{A}_d \le \tau_{\text{BF16}}$ are considered moderately critical and quantized to UINT4.
    \item \textbf{Low-Precision (UINT2):} Channels with $\mathcal{A}_d \le \tau_{\text{UINT4}}$ are deemed non-critical and compressed to UINT2.
\end{itemize}
The threshold determination process is detailed in Appendix~\ref{app:search_process}.

The overall workflow of MixKVQ is illustrated in Figure~\ref{fig:method}. 
We apply per-channel quantization to the Key cache and per-token quantization to the Value cache.
A buffer of size $R$ stores tokens in full precision until the buffer size is reached, after which they are
group quantized.
For the Key cache, the parameters $\mathcal{I}_d$ and $\mathcal{S}_d$ are updated periodically (every $R$ tokens). 
During this update, $\mathcal{I}_d$ is derived by averaging the absolute magnitudes of Query activations within the current window. 
Conversely, the Value cache undergoes uniform 2-bit per-token quantization.
Specifically, we provide the detail for MixKVQ when calculating the attention output in the prefill and decoding phases in Appendix~\ref{app: implement}.

%% file: latex/sections/5_experiment.tex
\section{Experiments}

\begin{table*}[!ht]
\centering
\caption{Performance comparison of BF16, KVquant, KIVI, KVTuner, RotateKV and MixKVQ on AIME 2024-2025, MATH 500, GPQA-Diamond, and LiveCodeBench datasets. 
% The maximum sequence length was set to 32768. 
% The maximum number of generation tokens is limited to 32,768. 
The best results are marked in bold.}
\vspace{-0.5em}
\resizebox{0.95\textwidth}{!}{
\begin{tabular}{llc|ccccc} 
\toprule
                                 &                                   &                                      & \multicolumn{4}{c}{\textbf{Accuracy (pass@1)}} \\ \cline{4-7}
\multirow{-2}{*}{\textbf{Model}} & \multirow{-2}{*}{\textbf{Method}} & \multirow{-2}{*}{\textbf{Bit-width}} & \textbf{AIME 2024-2025} & \textbf{MATH 500} & \textbf{GPQA-Diamond} & \textbf{LiveCodeBench} & \multirow{-2}{*}{\textbf{Avg.}} \\ \midrule
                                 & -                                  & BF16                                & 41.67 & 89.80 & 48.99 & 34.62 & 53.77 \\ \cline{2-8}
                                 & \multirow{-0.5}{*}{KIVI}           & KV4                                 & 35.00 & 86.60 & 46.46 & 28.57 & 49.16 \\
                                 &                                    & KV2                                 & 31.67 & 85.80 & 37.37 & 17.58 & 43.11 \\ \cline{2-8}
                                 & \multirow{-0.5}{*}{KVQuant}        & KV4                                 & 36.67 & 87.40 & 44.95 & 27.47 & 49.12 \\
                                 &                                    & KV2                                 & 0 & 4.20 & 21.72 & 12.64 & 9.64 \\ \cline{2-8}
                                 & RotateKV                           & KV4                                 & 36.67 & 88.00 & 42.93 & 30.22 & 49.45 \\ \cline{2-8}
\multirow{-6}{*}{\parbox{3.5cm}{\centering DeepSeek-R1- \\ Distill-Llama-8B}} 
                                 & \multicolumn{2}{l|}{KVTuner-C3.25}                                       & 30.00 & 87.80 & 46.96 & 31.32 & 49.02 \\ \cline{2-8}
                                 & \multicolumn{2}{>{\columncolor{gray!20}}l|}{MixKVQ-C2.7}                & \cellcolor{gray!20}\textbf{40.00} & \cellcolor{gray!20}\textbf{88.20} & \cellcolor{gray!20}\textbf{47.47} & \cellcolor{gray!20}\textbf{31.87} & \cellcolor{gray!20}\textbf{51.89} \\ \midrule
                                 & -                                  & BF16                                & 61.67 & 93.60 & 58.59 & 45.05 & 64.73 \\ \cline{2-8}
                                 & \multirow{-0.5}{*}{KIVI}           & KV4                                 & 55.00 & 92.20 & 56.06 & 42.31 & 61.39 \\
                                 &                                    & KV2                                 & 50.00 & 87.80 & 52.53 & 31.32 & 55.41 \\ \cline{2-8}
                                 & \multirow{-0.5}{*}{KVQuant}        & KV4                                 & 53.33 & 90.40 & 54.58 & 41.76 & 60.02 \\
                                 &                                    & KV2                                 & 0 & 11.00 & 22.22 & 14.29 & 11.88 \\ \cline{2-8}
                                 & RotateKV                           & KV4                                 & 56.67 & 92.40 & 55.56 & 42.31 & 61.73 \\ \cline{2-8} 
\multirow{-6}{*}{\parbox{3.5cm}{\centering DeepSeek-R1- \\ Distill-Qwen-14B}} 
                                 & \multicolumn{2}{l|}{KVTuner-C2.90}                                       & 41.67 & 89.80 & 50.00 & 40.11 & 55.40  \\ \cline{2-8}
                                 & \multicolumn{2}{>{\columncolor{gray!20}}l|}{MixKVQ-C2.3}                & \cellcolor{gray!20}\textbf{60.00} & \cellcolor{gray!20}\textbf{92.40} & \cellcolor{gray!20}\textbf{56.57} & \cellcolor{gray!20}\textbf{43.41} & \cellcolor{gray!20}\textbf{63.10} \\ \midrule
                                 & -                                  & BF16                                & 66.67 & 94.80 & 62.63 & 47.25 & 67.84 \\ \cline{2-8}
                                 & \multirow{-0.5}{*}{KIVI}           & KV4                                 & 58.33 & 93.40 & 59.09 & 44.51 & 63.83 \\
                                 &                                    & KV2                                 & 51.67 & 89.80 & 54.54 & 39.56 & 58.89 \\ \cline{2-8}
                                 & \multirow{-0.5}{*}{KVQuant}        & KV4                                 & 60.00 & 92.60 & 58.08 & 43.96 & 63.66 \\
                                 &                                    & KV2                                 & 0 & 42.80 & 30.81 & 20.33 & 23.48 \\ \cline{2-8}
                                 & RotateKV                           & KV4                                 & 61.67 & 93.40 & 60.10 & 42.86 & 64.51 \\ \cline{2-8}
\multirow{-6}{*}{\parbox{3.5cm}{\centering DeepSeek-R1- \\ Distill-Qwen-32B}} 
                                 & \multicolumn{2}{l|}{KVTuner-C2.91}                                       & 51.67 & 91.40 & 59.60 & 41.76 & 61.11 \\ \cline{2-8}
                                 & \multicolumn{2}{>{\columncolor{gray!20}}l|}{MixKVQ-C2.3}                & \cellcolor{gray!20}\textbf{65.00} & \cellcolor{gray!20}\textbf{94.00} & \cellcolor{gray!20}\textbf{60.10} & \cellcolor{gray!20}\textbf{45.05} & \cellcolor{gray!20}\textbf{66.04} \\ \bottomrule
\end{tabular}
}
\label{lab:reasoningbench}
\vspace{-1.0em}
\end{table*}

\subsection{Experimental Setup}
\paragraph{Models.}
We evaluate MixKVQ on four models including on the Deepseek-R1-Distill \cite{DeepSeekR1} series.
We choose Deepseek-R1-Distill-Qwen-14B/32B \cite{Qwen25} and Deepseek-R1-Distill-LLaMA-8B \cite{Llama3}.
All experiments are conducted on a single NVIDIA A800 GPU (80GB).

\paragraph{Tasks.}
We evaluate MixKVQ on four reasoning benchmarks: (1) AIME 2024-2025~\cite{AIME2024, AIME2025} and MATH-500~\cite{math500} for mathematical reasoning; (2) GPQA-Diamond~\cite{gpqa} for graduate-level scientific reasoning; and (3) LiveCodeBench~\cite{livecodebench} for code generation, specifically using the subset from January 1, 2025 to April 6, 2025.

\paragraph{Baseline.}
We compare our method with KVquant \cite{kvquant}, KIVI \cite{kivi}, KVTuner \cite{KVTuner}, RotateKV \cite{RotateKV} and BF16 baseline.
For all reasoning evaluations, we employ a sampling temperature of 0.6 and a top-$p$ of 0.95. 
To ensure a fair comparison, we standardize the group size at $G=32$ and the residual length at $R=128$ across all quantization methods, including our own.

\subsection{Main Results}
Table~\ref{lab:reasoningbench} presents the performance of MixKVQ on the AIME 2024-2025, MATH 500, GPQA-Diamond, and LiveCodeBench datasets. 
The results demonstrate that MixKVQ consistently outperforms existing methods across all evaluated models. 
For instance, on Qwen-32B model, MixKVQ attains an average accuracy of 66.04\%, closely trailing the BF16 baseline of 67.84\%.
In contrast, the 4-bit baseline RotateKV achieves an average accuracy of 64.51\% (a decrease of 3.33\%), while the 2-bit baseline KIVI drops to 58.89\% (a significant decline of 8.95\%). 
This indicates that MixKVQ can effectively mitigates the quantization error that typically impairs reasoning capabilities.

\begin{table*}[t]
    \caption{Performance comparison of BF16, KVQuant, KIVI, SKVQ, RotateKV and our MixKVQ on LongBench datasets. 
    The best results within each model group are marked in bold.}
    \vspace{-0.5em}
    \centering
    \resizebox{\textwidth}{!}{ 
    \begin{tabular}{lc|cccccccccc}
    \toprule
                     &                    & \multicolumn{2}{c}{\textbf{Single-document QA}}   & \multicolumn{2}{c}{\textbf{Summarization}} & \multicolumn{3}{c}{\textbf{Few-shot Learning}} & \multicolumn{2}{c}{\textbf{Code}} \\ \cmidrule(lr){3-4}\cmidrule(lr){5-6}\cmidrule(lr){7-9}\cmidrule(lr){10-11}
     \multirow{-2}{*}{\textbf{Method}} & \multirow{-2}{*}{\textbf{Bit-width}} & \textbf{Qasper} & \textbf{MultiFieldQA} & \textbf{QMSum} & \textbf{MultiNews} & \textbf{TREC} & \textbf{TriviaQA} & \textbf{SAMSum} & \textbf{LCC} & \textbf{RepoBench-P} & \multirow{-2}{*}{\textbf{Avg. $\uparrow$}} \\
    \midrule
    \multicolumn{12}{c}{\textbf{Mistral-7B-Instruct-v0.3}} \\
    \midrule
    - & BF16 & 41.58 & 55.23 & 25.75 & 27.82 & 76.00 & 88.59 & 47.35 & 60.53 & 61.68 & 53.84  \\
    \cline{1-12}
    \multirow{2}{*}{KVQuant} & KV4  & 40.56 & 53.41 & 24.72 & 26.79 & 75.00 & 88.21 & 46.97 & 58.08 & 58.13 & 52.43  \\
                             & KV2  & 37.88 & 49.27 & 22.16 & 25.89 & 70.00 & 86.53 & 45.38 & 56.42 & 55.94 & 49.94  \\
    \cline{1-12}
    \multirow{2}{*}{KIVI} & KV4 &  40.87 & 53.68 & 23.90 & 26.94 & 74.00 & 88.34 & 47.54 & 58.64 & 60.81 & 52.75  \\
                          & KV2 & 38.56 & 50.47 & 22.48 & 26.65 & 72.00 & 87.13 & 47.18 & 57.15 & 58.29 & 51.10  \\
    \cline{1-12}
    \multirow{2}{*}{SKVQ} & KV4 & 40.95 & 53.36 & 22.04 & 26.57 & 76.00 & \textbf{89.32} & \textbf{47.65} & 58.85 & 56.58 & 52.37  \\
                          & KV2 & 32.64 & 46.57 & 22.20 & 26.46 & 76.00 & 87.52 & 46.09 & 57.90 & 54.16 & 49.95  \\
    \cline{1-12}
    \multirow{2}{*}{RotateKV} & KV4 & 40.64 & 53.59 & 22.90 & 25.09 & 76.00 & 88.16 & 44.08 & 49.22 & 47.92 & 49.73 \\
                              & KV2 & 16.69 & 27.71 & 18.27 & 24.88 & 42.00 & 48.08 & 22.20 & 28.40 & 31.46 & 28.85 \\

    \cline{1-12}
    \multicolumn{2}{>{\columncolor{gray!20}}l|}{MixKVQ C2.7} & \cellcolor[HTML]{EFEFEF}\textbf{41.19} & \cellcolor[HTML]{EFEFEF}\textbf{54.52} & \cellcolor[HTML]{EFEFEF}\textbf{26.23} & \cellcolor[HTML]{EFEFEF}\textbf{27.66} & \cellcolor[HTML]{EFEFEF}\textbf{76.00} & \cellcolor[HTML]{EFEFEF}{88.59} & \cellcolor[HTML]{EFEFEF}{46.89} & \cellcolor[HTML]{EFEFEF}\textbf{60.44} & \cellcolor[HTML]{EFEFEF}\textbf{61.60} & \cellcolor[HTML]{EFEFEF}\textbf{53.68} \\
    \midrule
    \multicolumn{12}{c}{\textbf{Llama-3.1-8B-Instruct}} \\
    \midrule
    - & BF16 & 45.53 & 56.37 & 25.36 & 27.19 & 72.50 & 91.65 & 43.62 & 65.10 & 58.65 & 54.00 \\
    \cline{1-12}
    \multirow{2}{*}{KVQuant} & KV4 & 44.03 & 51.67 & 25.25 & 26.87 & 71.00 & 91.03 & 43.61 & 62.15 & 55.13 & 52.30  \\
                             & KV2 & 36.51 & 41.85 & 22.94 & 25.82 & 65.00 & 86.27 & 41.94 & 59.88 & 50.87 & 47.90  \\
    \cline{1-12}
    \multirow{2}{*}{KIVI} & KV4 & 44.42 & 52.10 & 25.28 & 26.86 & 72.50 & 91.42 & \textbf{44.07} & 62.45 & 55.04 & 52.68  \\
                          & KV2 & 41.02 & 45.86 & 24.57 & 26.79 & 71.50 & 91.14 & 43.41 & 60.55 & 51.64 & 50.72  \\
    \cline{1-12}
    \multirow{2}{*}{SKVQ} & KV4 & 44.21 & 53.96& \textbf{25.34} & 26.79 & 72.50 & 91.14 & 42.38 & 62.44 & 54.13 & 52.54 \\
                          & KV2 & 35.28 & 46.27 & 23.61 & 26.47 & 72.50 & 91.21 & 42.96 & 60.63 & 50.63 & 49.95 \\
    \cline{1-12}
    \multirow{2}{*}{RotateKV} & KV4 & 44.32 & 53.92 & 25.00 & 27.03 & 71.50  & 90.84 & 43.12 & 63.88 & 56.39 & 52.89 \\
                              & KV2 & 10.95 & 22.64 & 17.50 & 20.37 & 21.00 & 9.84 & 8.18 & 30.29 & 29.20 & 18.89 \\
    \cline{1-12}
    \multicolumn{2}{>{\columncolor{gray!20}}l|}{MixKVQ C2.7} & \cellcolor[HTML]{EFEFEF}\textbf{45.32} & \cellcolor[HTML]{EFEFEF}\textbf{55.40} & \cellcolor[HTML]{EFEFEF}{25.29} & \cellcolor[HTML]{EFEFEF}\textbf{27.20} & \cellcolor[HTML]{EFEFEF}\textbf{72.50} & \cellcolor[HTML]{EFEFEF}\textbf{91.78} & \cellcolor[HTML]{EFEFEF}{43.68} & \cellcolor[HTML]{EFEFEF}\textbf{64.96} & \cellcolor[HTML]{EFEFEF}\textbf{57.26} & \cellcolor[HTML]{EFEFEF}\textbf{53.71} \\
    \bottomrule
    \end{tabular}
    }
    \label{tab:longbench}
    \vspace{-1.0em}
\end{table*}

\begin{figure}[h!]
    \begin{center}
    \includegraphics[width=0.95\linewidth]{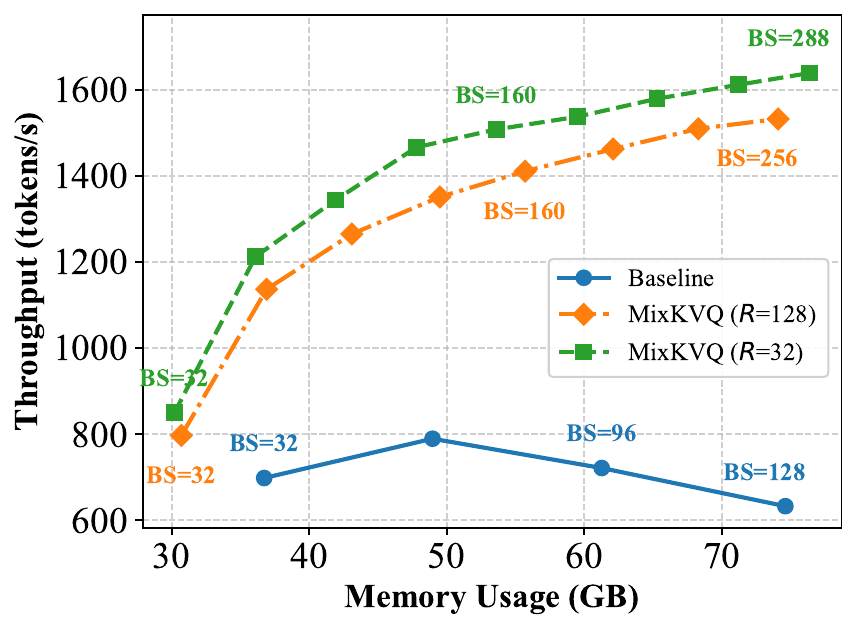}
    \end{center}
    \vspace{-1.0em}
    \caption{Memory usage and throughput comparison between MixKVQ and 16 bit baseline.}
    \label{fig: mem_speed}
    \vspace{-1.5em}
\end{figure}

\subsection{Long Context Generation Accuracy}
Table \ref{tab:longbench} presents the evaluation on LongBench \cite{longbench} for Mistral-7B and Llama-3.1. 
Compared to BF16 and competitive baselines, MixKVQ pushes the effective bit-width down to 2.70 bits with negligible performance degradation. 
This result highlights the efficacy of our mixed-precision strategy, which surpasses quantization baselines in delivering both high fidelity of model performance and memory efficiency.

\subsection{Efficiency Comparison}
We assess the runtime efficiency of MixKVQ using workloads synthesized from ShareGPT~\footnote{\url{https://sharegpt.com/}}, strictly following the vLLM evaluation setting~\cite{vllm}. 
% ($l_{\text{prompt}}=161$, $l_{\text{gen}}=338$). 
We push the batch size to memory saturation and compare the throughput and peak memory usage of MixKVQ (residual lengths 32/128) with the FP16 baseline on Llama2-13B-chat.
As illustrated in Figure~\ref{fig: mem_speed}, with similar maximum memory usage, MixKVQ enables up to $\mathbf{2.25\times}$ larger batch size and gives $\mathbf{2.63\times \sim 2.81\times}$ larger throughput. 
These performance gains are expected to amplify with extended sequence lengths. 
Furthermore, we plan to integrate MixKVQ into high-performance serving frameworks like vLLM in future work, which we anticipate will unlock substantial additional throughput gains.

Ablation studies investigating the sensitivity to hyperparameters (group size $G$, residual length $R$) and the contribution of the query-aware component ($\mathcal{I}_d$) are provided in Appendix~\ref{app: ablation}.

%% file: latex/sections/6_conclusion.tex
\section{Conclusion}
In this paper, we propose MixKVQ, a plug-and-play KV cache quantization algorithm without the need for any tuning. 
MixKVQ allocates the precision for a key channel depending on two factors: its intrinsic quantization difficulty and its dynamic relevance to the query.
Based on this strategic, MixKVQ quantizes key cache per-channel mixed precision and quantizes value cache per-token.
The evaluation shows that MixKVQ achieves optimal performance under extreme low-bit widths in complex reasoning tasks.

%% file: latex/sections/7_limitationsAndRisks.tex
\section*{Limitations}
While MixKVQ achieves substantial KV cache compression via ultra-low-bit quantization, its computational overhead during tensor transformation remains non-negligible despite our GPU-optimized mitigation strategies. 
This limitation suggests opportunities to enhance runtime efficiency through deeper integration with LLM inference accelerators like vLLM.
Additionally, our study does not encompass all attention mechanisms; specifically, we exclude Multi-Head Latent Attention (MLA), which differs significantly from the widely adopted Group-Query Attention (GQA).
Furthermore, our latency analysis focuses primarily on memory-bound generation stages, leaving the computational bottlenecks in prompt processing phases insufficiently explored, particularly during batch compression operations across multiple key-value sequences. 

%% file: latex/sections/8_appendix.tex
\section{Derivation of Quantization Error Bound}
\label{app:quant_error_proof}

In this section, we provide a formal derivation for the quantization error bound mentioned in Section~\ref{sec:quant} (assuming your main text is in a section labeled like this).

Recall the asymmetric quantization and dequantization operations defined as:
\begin{align}
    q_i &= \text{round}\left( \frac{x_i - z}{s} \right), \\
    \tilde{x}_i &= q_i \cdot s + z,
\end{align}
where $x_i \in \boldsymbol{X}$ is the original value, $q_i$ is the quantized integer, $\tilde{x}_i$ is the dequantized approximation, $z$ is the zero-point, and $s$ is the scaling factor ($s > 0$).

We aim to find the upper bound of the absolute quantization error, denoted as $|x_i - \tilde{x}_i|$.
By substituting the definition of $\tilde{x}_i$ into the error term, we have:
\begin{equation}
    |x_i - \tilde{x}_i| = |x_i - (q_i \cdot s + z)|.
\end{equation}
Rearranging the terms inside the absolute value operator:
\begin{equation}
    |x_i - \tilde{x}_i| = |(x_i - z) - q_i \cdot s|.
\end{equation}
We can factor out the scaling factor $s$:
\begin{equation}
    |x_i - \tilde{x}_i| = \left| s \cdot \left( \frac{x_i - z}{s} - q_i \right) \right|.
\end{equation}
Since $s$ is a positive scalar derived from the dynamic range, we can move it outside the absolute value:
\begin{equation}
    |x_i - \tilde{x}_i| = s \cdot \left| \frac{x_i - z}{s} - q_i \right|.
\end{equation}
Let $y_i = \frac{x_i - z}{s}$. By definition, $q_i = \text{round}(y_i)$. The rounding operation $\text{round}(\cdot)$ maps a real number to the nearest integer. A fundamental property of this operation is that the absolute difference between a real number and its nearest integer is at most $0.5$:
\begin{equation}
    |y_i - \text{round}(y_i)| \le \frac{1}{2}.
\end{equation}
Substituting this inequality back into our error equation:
\begin{align}
    |x_i - \tilde{x}_i| &= s \cdot |y_i - q_i| \nonumber \\
    &\le s \cdot \frac{1}{2} \nonumber \\
    &= \frac{s}{2}.
\end{align}
Thus, the quantization error for any element $x_i$ is bounded by half the scaling factor $s$.

\section{Failure Case Analysis of KVTuner}
\label{app:kvtuner_failure}
\begin{figure*}[t!]
\begin{center}
    \includegraphics[width=1.0\textwidth]{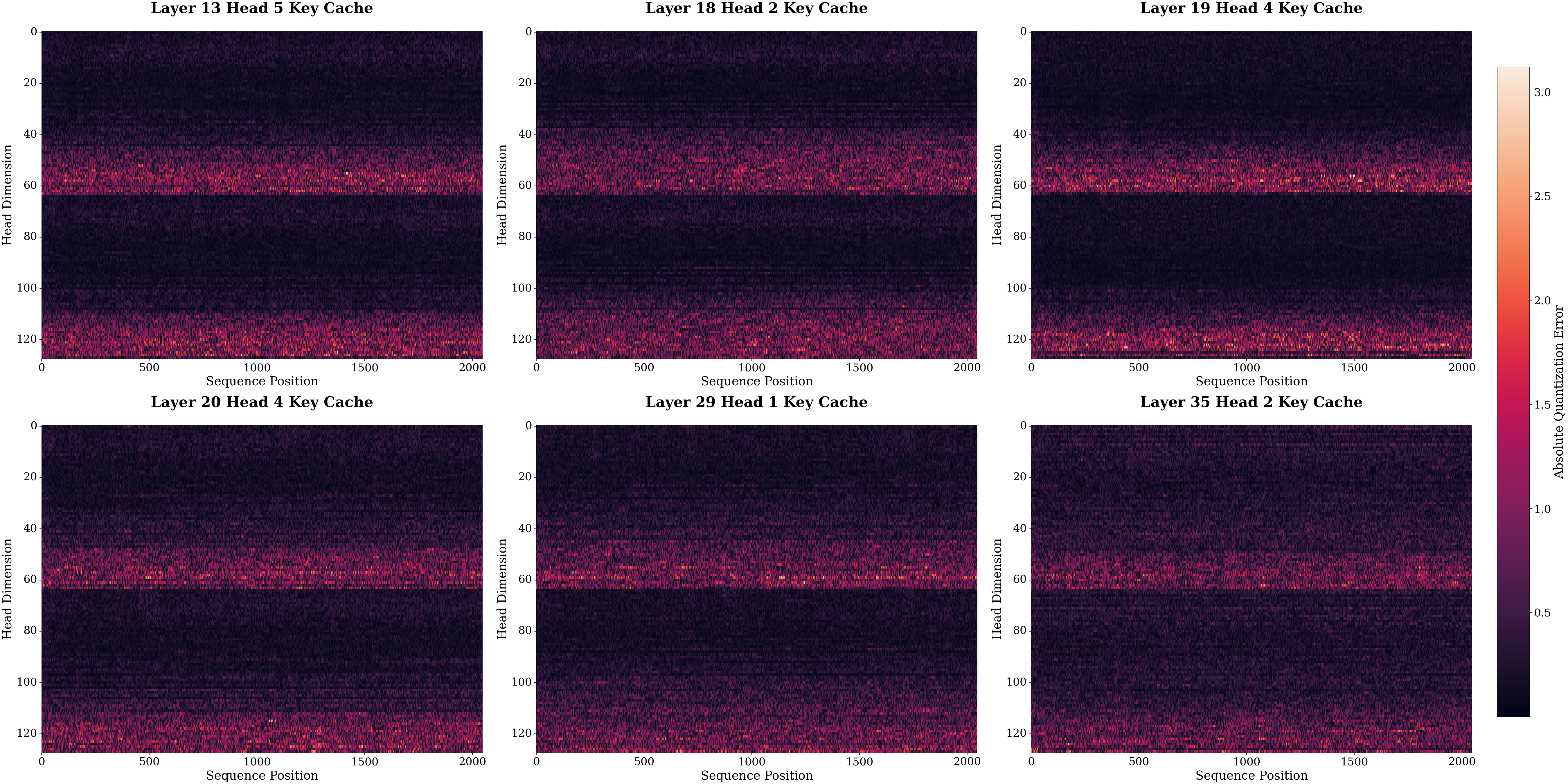}
    \end{center}
    \caption{Failure case analysis of the KVTuner method. KVTuner statically identifies many layers as non-critical and applies a uniform, aggressive K2V2 (2-bit key, 2-bit value) quantization policy to them. However, as visualized in the heatmaps, even these targeted layers contain specific dimensions with outlier features that resist effective quantization.
    }
    \label{fig:faliure_example}
    % \vspace{-1.0em}
\end{figure*}

KVTuner operates by leveraging calibration data to statically identify entire layers deemed ``less critical.'' 
To meet a target memory budget, these designated layers are then subjected to aggressive K2V2 quantization. 
However, this layer-level approach has a fundamental limitation. 
As illustrated in Figure~\ref{fig:faliure_example}, even within these so-called non-critical layers, there exist specific dimensions that are difficult to quantize due to outlier values. 
Uniformly applying an aggressive quantization policy to the entire layer results in significant information loss in these critical dimensions, introducing substantial errors that degrade the model's performance on complex, multi-step reasoning tasks.

\begin{figure*}[t!]
\begin{center}
    \includegraphics[width=1.0\textwidth]{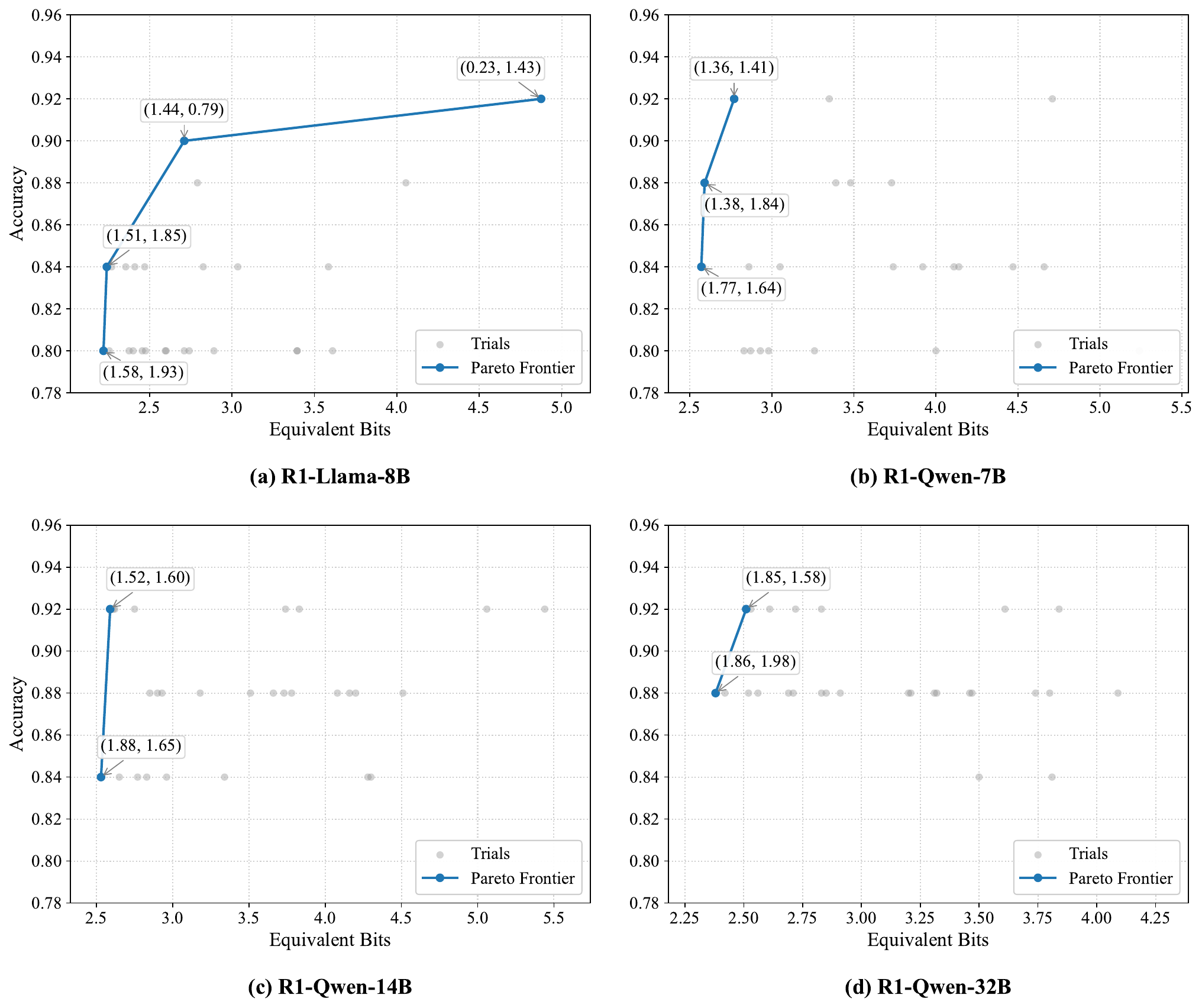}
    \end{center}
    \caption{Pareto frontier of different models with the MixKVQ quantization mode on the 25 data slices of the GSM8K dataset.
    }
    \label{fig:pareto}
    \vspace{-1.0em}
\end{figure*}

\section{Threshold Search Strategy}
\label{app:search_process}
This section details the procedure for determining the importance thresholds, $\tau_{\text{BF16}}$ and $\tau_{\text{INT4}}$, which govern the mixed-precision allocation in the KV cache.
We formulate the threshold selection as a dual-objective optimization problem.
This framework facilitates the joint optimization of both hyperparameters to effectively navigate the trade-off between memory compression and performance fidelity.

\subsection{Optimization Setup}
We employ the \textsc{Optuna} framework to conduct a joint search for $\tau_{\text{BF16}}$ and $\tau_{\text{INT4}}$ within the search space $[0.1, 2.0]$.
The optimization aims to satisfy two conflicting objectives simultaneously:
\begin{enumerate}
    \item \textbf{Maximize Accuracy:} We evaluate the quantized model on the GSM8K benchmark. This metric ensures that the quantization process preserves the model's complex chain-of-thought reasoning abilities.
    \item \textbf{Minimize Effective Bit-width ($B_{\text{eff}}$):} We aim to minimize the average bit-width per parameter, defined as:
    \begin{equation}
        B_{\text{eff}} = \frac{1}{N} \sum_{i=1}^{N} b_i(\tau_{\text{BF16}}, \tau_{\text{INT4}})
    \end{equation}
    where $N$ is the total number of channels, and $b_i \in \{16, 4, 2\}$ denotes the bit-width assigned to the $i$-th channel based on the selected thresholds.
\end{enumerate}

\subsection{Selection Process}
We utilize the Tree-structured Parzen Estimator (TPE) sampler to efficiently explore the hyperparameter space.
By executing 30 trials, we construct a Pareto frontier that visualizes the optimal trade-offs between accuracy and memory efficiency.
From this frontier, we select the configuration that yields the highest accuracy while maintaining the effective bit-width below a strict constraint.

As illustrated in Figure~\ref{fig:pareto}, the optimal thresholds vary across architectures, reflecting their differing sensitivities to quantization.
For the \textbf{R1-Llama-8B}, we select thresholds $(\tau_{\text{BF16}}, \tau_{\text{INT4}}) = (1.44, 0.79)$, resulting in an effective bit-width of 2.7 bits.
In contrast, the \textbf{R1-Qwen-7B} model exhibits higher sensitivity, requiring more conservative thresholds of $(0.63, 0.41)$ which yield 3.4 bits.
Larger models demonstrate greater robustness to compression.
specifically, \textbf{R1-Qwen-14B} and \textbf{R1-Qwen-32B} both achieve an effective bit-width of 2.3 bits with thresholds of $(1.52, 1.60)$ and $(1.85, 1.58)$, respectively.

\section{Detailed Implementations}
\label{app: implement}
This section details the implementation of the MixKVQ algorithm proposed in Section~\ref{method}. 

To ensure compatibility with modern Large Language Model architectures, our quantization granularity is strictly aligned with the attention mechanism. 
For models employing Grouped Query Attention (GQA), importance scores are computed at the KV head group level. 
Specifically, we aggregate query magnitudes from all query heads corresponding to a shared KV head to determine the appropriate quantization precision.

\paragraph{Storage Layout}
The KV cache is organized into three distinct components to balance memory efficiency and retrieval precision:
\begin{itemize}
    \item \textbf{Quantized Storage ($Q(\mX_{K/V})$):} This component stores the majority of historical states. 
    Data is packed into low-bit contiguous tensors (e.g., UINT4 or UINT2) to maximize memory throughput.
    \item \textbf{Sparse Outlier Storage ($\mX_{K_{BF16}}$):} Salient channels identified by the importance thresholds are retained in full BF16 precision. These outliers are managed using sparse formats to minimize storage overhead.
    \item \textbf{High-Precision Residual Buffer ($\mX_{R}$):} A compact buffer maintained in full precision to temporarily store the most recent tokens before quantization.
\end{itemize}

\subsection{Amortized Scheduling via Lazy Updates}
To mitigate the computational overhead associated with frequent re-quantization, we employ the lazy update strategy controlled by a residual length hyperparameter $R$.

During the decoding phase, newly generated Key and Value states are initially appended to the full-precision residual buffer $\mX_{R}$. 
Consequently, quantization operations are not executed at every generation step. 
Computationally intensive tasks, including channel selection, outlier extraction, and bit-packing, are triggered exclusively when the buffer length reaches $R$.
Upon triggering, the accumulated block is processed via the \FuncSty{KeyQuant} procedure, partitioned into quantized and outlier components, and subsequently merged into the main cache. 
The buffer is then reset. 
This lazy update mechanism serves a critical dual purpose: it not only amortizes the computational overhead across $R$ decoding steps but also functions as a temporal stabilization window. 
Since channel salience ($\mathcal{I}_d$) often exhibits transient volatility due to local attention patterns on recent tokens, immediate quantization risks assigning precision based on unstable statistics. 
By isolating these volatile states within $\mX_{R}$, we defer the quantization decision until tokens transition out of the local window.

\subsection{Efficient Online Saliency Estimation}
Calculating the cumulative importance of KV channels typically necessitates scanning the entire query history, which is computationally prohibitive. 
To address this, we maintain a running accumulator of query magnitudes within the cache. 
At each decoding step, the magnitude of the current query token is added to this accumulator. 

Furthermore, for architectures incorporating Rotary Positional Embeddings (RoPE), saliency evaluation is performed post-transformation. 
Specifically, importance scores are calculated after applying rotary embeddings to the query and key states. 
This ensures that the metric accurately reflects the attention distribution in the rotated space.

\begin{table}
\centering
\small
\caption{Ablation study of MixKVQ by changing group size $G$ and residual length $R$.}
% \vspace{-.5em}
\label{tab:abl}
\begin{tabular}{lcc} 
\toprule
\textbf{Model}                    & \textbf{Group Size} & \textbf{PPL$\downarrow$}  \\ 
\midrule
\multirow{3}{*}{Llama2-13B-chat}  & 32                  & 7.05 \\
                                  & 64                  & 7.10 \\
                                  & 128                 & 7.12\\
\bottomrule
\toprule
\textbf{Model}                    & \textbf{Residual Length} & \textbf{PPL$\downarrow$}  \\ 
\midrule
 \multirow{5}{*}{Llama2-13B-chat} & 32                       & 7.09 \\
                                  & 64                       & 7.04 \\
                                  & 96                       & 7.03 \\
                                  & 128                      & 7.05 \\
                                  & 256                      & 7.03 \\
\bottomrule
\end{tabular}
\vspace{-0.5em}
\end{table}

\begin{table}
\centering
\small
\caption{Ablation study of MixKVQ by isolating query-aware component.}
% \vspace{-.5em}
\label{tab:abl_query}
\begin{tabular}{lcc} 
\toprule
\textbf{Model}                & \textbf{Method} & \textbf{AIME 2024-2025}  \\ 
\midrule
\multirow{2}{*}{R1-Qwen-14B}  & error-only      & 53.33 \\
                              & MixKVQ          & 60.00 \\ \cline{1-3}
\multirow{2}{*}{R1-Llama-8B}  & error-only      & 33.33 \\
                              & MixKVQ          & 40.00 \\
\bottomrule
\end{tabular}
\vspace{-0.5em}
\end{table} 

\section{Ablation Study}
\label{app: ablation}
Specifically, we investigate the sensitivity of model performance to the quantization group size $G$ and residual length $R$, and further isolate the contribution of the query-aware saliency term ($\mathcal{I}_d$) against error-only baselines..

\paragraph{The effect of group size.}
We fix the residual length at 128 and the sink length at 32.
We vary the group sizes to 32, 64, and 128. 
From Table \ref{tab:abl}, we observe that PPL decreases with group sizes. Note the zero-point and the scaling factor are calculated according to this group size. The choice of group size will greatly impact the KV cache compression effect under a long input.

\paragraph{The effect of residual length.}
We fix the group size at 32 and the sink length at 32.
We vary the residual length across 32, 64, 96, 128, and 256. 
As shown in Table~\ref{tab:abl}, there is no consistent pattern between residual lengths and model accuracy.
Although no significant differences were observed among the tested residual lengths {32, 64, 96, 128, 256}, a sufficiently large residual length remains crucial, offering considerable performance boosts for difficult tasks.

\paragraph{Necessity of the query-aware component.} 
To validate the efficacy of our composite metric $\mathcal{A}_d = \mathcal{I}_d \times \mathcal{S}_d$, we conducted a comparative analysis against a Error-only baseline, which determines precision solely based on channel magnitude ( $\mathcal{A}_d = \mathcal{S}_d$). 
We evaluated the variant on the AIME 2024-2025 reasoning benchmark. 
As presented in Table~\ref{tab:abl_query}, the full MixKVQ method outperforms the error-only baseline. 
This result confirms that relying exclusively on quantization error minimization is insufficient, explicitly incorporating query-dependent importance ($\mathcal{I}_d$) is indispensable for preserving model fidelity during complex reasoning tasks.

\begin{table}[htbp!]
    \centering
    \caption{Per-layer time breakdown (\%) and call rates across decode steps.}
    \label{tab:overhead_analysis}
    \resizebox{\columnwidth}{!}{%
        \begin{tabular}{lcc}
            \toprule
            \textbf{Operation} & \textbf{Time Breakdown (\%)} & \textbf{\# of Calls (\%)} \\
            \midrule
            Channel Selection & 2.17  & 3.13 \\
            Attention         & 64.62 & 100 \\
            MLP               & 33.21 & 100 \\
            \bottomrule
        \end{tabular}%
    }
\end{table}

\section{Computation Overhead}
\label{app: overhead}
In Table \ref{tab:overhead_analysis}, we report operation-level breakdowns for R1-Qwen-7B. 
While Channel Selection comprise 2.17\% of per-layer execution, which enables a KV cache memory savings of over 79\% (compressing from 16-bit to 3.4-bit).

\begin{table*}[htbp!]
\centering
\caption{Performance comparison of BF16, KVquant, KIVI, KVTuner, RotateKV and MixKVQ on AIME 2024-2025, MATH 500 and GPQA-Diamond datasets. 
The maximum number of generation tokens is limited to 32,768. 
The best results are marked in bold.}
\resizebox{\textwidth}{!}{
\begin{tabular}{llc|ccccc} 
\toprule
                                 &                                   &                                      & \multicolumn{4}{c}{\textbf{Accuracy (pass@1)}} \\ \cline{4-7}
\multirow{-2}{*}{\textbf{Model}} & \multirow{-2}{*}{\textbf{Method}} & \multirow{-2}{*}{\textbf{Bit-width}} & \textbf{AIME 2024-2025} & \textbf{MATH 500} & \textbf{GPQA-Diamond} & \textbf{LiveCodeBench} & \multirow{-2}{*}{\textbf{Avg.}} \\ \midrule
                                 & -                                  & BF16                                & 46.67 & 89.00 & 49.49 & 23.08 & 52.06 \\ \cline{2-8}
                                 & \multirow{-0.5}{*}{KIVI}           & KV4                                 & 36.67 & 87.00 & 44.94 & \textbf{20.88} & 47.37 \\
                                 &                                    & KV2                                 & 30.00 & 84.60 & 35.35 & 16.48 & 41.61 \\ \cline{2-8}
                                 & \multirow{-0.5}{*}{KVQuant}        & KV4                                 & 33.33 & 87.40 & 41.41 & 19.44 & 45.40 \\
                                 &                                    & KV2                                 & 1.67  & 37.80 & 25.76 & 5.49 & 17.68 \\ \cline{2-8}
                                 & RotateKV                           & KV4                                 & 33.33 & 86.40 & 42.93 & 18.33 & 45.25 \\ \cline{2-8}
\multirow{-7}{*}{\parbox{3.5cm}{\centering DeepSeek-R1- \\ Distill-Qwen-7B}} 
                                 & \multicolumn{2}{l|}{KVTuner-C3.92}                                       & 31.67 & 86.00 & 39.90 & 19.23 & 44.20 \\ \cline{2-8}
                                 & \multicolumn{2}{>{\columncolor{gray!20}}l|}{MixKVQ-C3.4}                & \cellcolor{gray!20}\textbf{40.00} & \cellcolor{gray!20}\textbf{88.20} & \cellcolor{gray!20}\textbf{45.45} & \cellcolor{gray!20}20.33 & \cellcolor{gray!20}\textbf{48.49} \\ \bottomrule
\end{tabular}
}
\label{lab:additional_reasoningbench}
\end{table*}

\section{Additional Experiments}
\label{app: additional}
In this section, we benchmark MixKVQ on DeepSeek-R1-Distill-Qwen-7B \cite{Qwen25}.